# A Novel Approach to Fast Image Filtering Algorithm of Infrared Images based on Intro Sort Algorithm

Kapil Kumar Gupta[1], M. Rizwan Beg[2], Jitendra Kumar Niranjan[3]

[1] Department of Computer Science & Engg., Integral University,
Lucknow, Uttar Pradesh, 226001, India

[2] Department of Computer Science & Engg., Integral University,
Lucknow, Uttar Pradesh, 226001, India

[3] Department of Computer Science & Engg, IMS Engineering College
Ghaziabad, Uttar Pradesh 201009, India

**Abstract**
In this study we investigate the fast image filtering algorithm based on Intro sort algorithm and fast noise reduction of infrared images. Main feature of the proposed approach is that no prior knowledge of noise required. It is developed based on Stefan-Boltzmann law and the Fourier law. We also investigate the fast noise reduction approach that has advantage of less computation load. In addition, it can retain edges, details, text information even if the size of the window increases. Intro sort algorithm begins with Quick sort and switches to heap sort when the recursion depth exceeds a level based on the number of elements being sorted. This approach has the advantage of fast noise reduction by reducing the comparison time. It also significantly speed up the noise reduction process and can apply to real-time image processing. This approach will extend the Infrared images applications for medicine and video conferencing.

## 1. Introduction

In Infrared images, impulse noise detection and removal is an important process as the images are corrupted by those noise because of transmission and acquisition. The main aim of the noise removal is to suppress the noise when preserving the edge information. Images and videos belong to the most important information carriers in today's world (e.g., traffic observations, surveillance systems, autonomous navigation, etc.). However, the images are likely to be corrupted by noise due to bad acquisition, transmission or recording. Such degradation negatively influences the performance of many image processing techniques and a preprocessing module to filter the images is often required.

The sensitive spectrum of an IR camera is about 3–5µm and 8–14µm. So, the IR images are robust under a wide-range of lighting conditions. However, the low signal-to-noise (S/N) ratio [1,2] is the inherent limitation of IR images that affect their quality and hinder their deployment. The low S/N ratio results in low signal and high noise that degrades the quality of IR images. This is significant for un-cooled IR camera, even though the un-cooled IR camera is much cheaper than the cooled one and more prevalently used to capture IR images in recent years. The high noise is caused by the IR sensors and read-out circuits of IR cameras, and the low IR signal detected by IR sensors is due to the bad atmospheric weather's degrading the IR signal radiating from objects. To enhance image quality and improve the adoption of IR-based applications, image preprocessing is necessary. Improvement in noise reduction is the crucial task of IR image preprocessing.

In this paper, a Fast Image Filtering approach to Infrared images is developed that is based on *Infrared imaging mechanism* to detect noise and median-based to remove noise with low computation load. It is performed without any prior knowledge about the IR image noise is necessary and any parameters must be preset. This property is quite different from some state-of-the-art noise reduction methods [5–8] which performance relies on one or more external heuristically preset parameter.

## 2. Previous Research

The standard median filter (MF) [3] has been prevalently used in for noise reduction of image preprocessing. However, there are two inherent limitations of the MF. The first is high computation load. The second is that it removes the thin lines and small objects of interest and blurs the details even at low noise densities, while the size of the filter window increasing. The later makes it worse when the objects of interest are with few pixels in IR images. The MF will consider the few pixel objects as noise and remove them. The weighted median filter and the center-weight median filter (CWMF) [3,4] which are modified the MF to alleviate the inherent limitations of MF at the expense of reduced noise removal performance. In addition, there are many methods [5–20] combine the MF with impulse detection have been proposed to remedy the MF's limitations. Their performances inherently rely on the performance of the impulse detector. Mean-based filters [21–24] are the alternative approach to remedy the limitations of MF. These filter usually exhibit good filter performance at the cost of increased computational complexity. In recent years, a number of literatures [25–28] proposed the impulse noise reduction based on fuzzy technologies. However all these literatures mentioned previously considered on visual images.

## 3. Fast noise reduction approach

There is a main difference between visual image and infrared image imaging mechanism. The visual sensor receives the visual light reflected from object's surface to image visual image. It needs an external light source to offer a sufficient light power. This approach is called active imaging mechanism. An object reflects the light power based on the texture, color, roughness and other factors of its surface. These factors induce the reflected light power irregularly. Median filter filters the impulse noise from visual images based on an assumption that signal pixels have high correspondence with their neighbor pixels inside a small area. This assumption is reasonable but not theoretical. Because of this, the images processed by median filter will possibly lose some thin lines, textures or details. On the other hand, the IR sensor receives the infrared emitted from objects themselves to image IR images. It does not need external light sources to offer the IR to illuminate objects. This approach is called passive imaging mechanism. The temperature distribute on object's surface monotonically. Thus, object's surface also emits infrared power to IR sensor monotonically. The gray-level of signal pixels on the same object surface will vary monotonically while noise pixels cannot do it. The proposed algorithm developed based on the infrared image imaging mechanism theoretically. The details are described in following.

### 3.1 Stephen-Boltzmann law

The imaging mechanism is derived based on the Stefan–Boltzmann law [29] (heat radiation law) and the Fourier law (heat conduction law). The heat radiation law is shown in Eq. (1):

$$PW = \epsilon \sigma T^4 \qquad (1)$$

where PW is the radiant emittance (W/cm2), $\epsilon$ is the emissivity, $\sigma$ is the Stefan–Boltzmann constant ($\approx 5.6705 \times 10^{-12}$ W/cm$^2$ K$^4$), T is the temperature (K) of the object surface. The intensity of IR radiation emitted by the objects in the range of 3–14 µm [30] is dependent on the emissivity of the object surface, the surface temperature, the air molecules, the humidity of the air, and the distance between the IR camera and the objects. The IR transmission spectrum for the atmosphere is about 3–5 µm and 8–14 µm [31], which means that the radiant emittance of the IR spectrum at 3– 5 lm and 8–14 lm has only minimum attenuation in the atmosphere. In Eq. (1), σ is a constant and e is also a constant for objects The temperature T of the object surface is the only variable that dominates the PW that significantly affects the thermal image contrast and quality.

### 3.2 Fourier law (heat conduction law)

the gray-level of pixels has a positive relationship with the temperature T of the object surface. Moreover, heat conduction affects the temperature distribution on the skin surface. Based on the Fourier law (heat conduction law (2)), the temperature gradient will vary over the surface of an object, and the direction of the temperature gradient changes slowly from high to low.

$$Q_1 = -kA (dT/dl), \qquad (2)$$

where $Q_1$ is the rate of heat flow through area A in the positive l direction, and the constant k is the thermal conductivity of the material. The heat conduction law means that the rate of heat flow is proportional to the area and the temperature gradient in a given direction. Thus, the temperature T of the object varies monotonically on a surface.

### 3.3 Noise detection

In order to explain how to perform the noise detection, we must define some parameters of image pixels. x and y represent the horizontal and vertical coordinates of a pixel, respectively. p(x, y) is the pixel with coordinates x and y. g(x, y) is the gray-level of the pixel p(x, y). $S_{xy}$ is the set that includes g(x, y) and its neighbor pixels. For example, a 3×3 window, $S_{xy}$ = {g(x -1,y - 1), g(x, y - 1), g(x + 1, y-1), g(x -1, y), g(x,y), g(x + 1, y), g(x -1, y + 1), g(x, y + 1), g(x + 1, y + 1)}, $g_m$ is the median gray-level of the $S_{xy}$, g(x, y) represents the gray-level of the central position pixel that will be processed, (s, t) denotes the coordinates of the pixels belonging to $S_{xy}$, and g(s, t) represents the gray- level of the pixels belonging to $S_{xy}$.

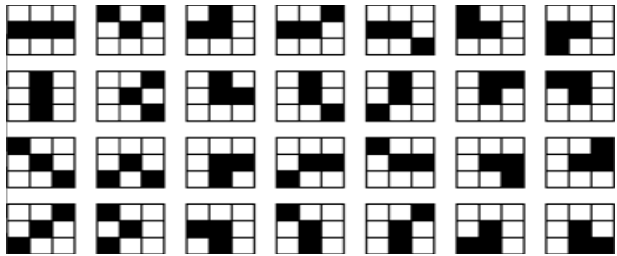

Fig. 1. Comparisons of central pixel to its two neighbors pixels in 3×3 Window

we propose the noise detection algorithm to find noisy pixels inside the filter window in IR images based on two steps. First, the noise detection method employs Eq. (3a) and (3b) to find the maximum and minimum gray-level inside the filter window. Next, we check the gray-level $g(x, y)$ with $g_{min}$ and $g_{max}$ to consider whether the pixel $p(x, y)$ is noise or signal. This process can be performed in the following steps.

Step 1. Determine the maximum and minimum gray-level inside the filter window.

$g_{max}$ = Arg max{g(s,t)}
        (s,t) ϵ $S_{xy}$             (3a)
$g_{min}$ = Arg min{g(s,t)}
        (s,t) ϵ $S_{xy}$             (3b)

Step 2. Based on the IR imaging mechanism, check $g(x, y)$ with gray-level $g_{min}$ and $g_{max}$ to determine whether the pixel $p(x, y)$ is noise or signal.
If $g(x, y) = g_{min}$ or $g_{max}$ then $p(x, y)$ is a noisy pixel
Otherwise $p(x, y)$ is considered as a signal pixel.

### 3.4 Noise removal based on Intro sort algorithm

In order to reduce noise, a pixel is considered as a noisy pixel which has to be removed from the IR image. According to the property of IR imaging mechanism, the pixel with median gray-level inside the window is adopted replacing the noisy pixels. In the proposed approach, the procedure to find out median gray-level is performed by the sort algorithm with a low computation complexity. In addition, it only processes the noisy pixels, but not the signal pixels.

Sorting is the main computation load of the noise removal. So, in order to speed up the noise removal, reducing the computation load is critical. This paper adopts a sort algorithm with low computation load. Sort algorithms are normally divided into two groups called internal sort and external sort. The former is suitable to small databases, and the latter is usually used on large databases. Hence, the number of pixels inside the filter window is small that may be 3 ×3, 5 ×5, 7 ×7, and so on. Based on the properties of the sort algorithm and the number of pixels sorted, we adopt the internal sort algorithm to sort the pixel gray-levels inside a filter window. The internal sort algorithm includes Bubble sort, Insertion sort, Selection sort, Shell sort, Quick sort, Heap sort, Radix sort, and so on. [32,33]. From an analysis of the properties of the versatile sort algorithms, a Intro sort with suitable parameters should run faster even than Quick sort or Heap sort. The complexity of Intro sort is described as Eq. (4).

Complexity of Intro sort  = O(n log n)        (4)
Where n is the number of data.

It begins with Quick sort and switches to Heap sort when the recursion depth exceeds a level based on the number of elements being sorted. It is the best of both worlds, with a worst-case O($n$ log $n$) runtime and practical performance comparable to Quick sort on typical data sets. Since both algorithms it uses are comparison sorts, it is a comparison sort too. Then the Intro sort algorithm is performed on bits to select the pixel with the median gray-level $g_m$ inside a filter window. The selected pixel with gray-level gm is utilized to replace the noisy pixel in the central position inside the filter window and the noise removal is accomplished. The Fast noise reduction approach does not act like the Median Filter in sorting each pixel in a whole Infrared image; it only processes that on noisy pixels. In general, there are far fewer noisy pixels than signal pixels. The FNR approach has to spend extra computation load to determine the maximum and minimum gray-level in the noise detection procedure.                (3a)

### 4. Experimental Results

4.1 Platform for evaluation

The platform utilized to evaluate the proposed approach includes a dual core CPU, the Intel Core 2 E6600 with clock rate 2.4 GHz and memory 1 Gbytes DDR2 667. The display card has GPU GeForce 7600 GT of NVIDIA Inc. and 256 MB memory. The program to simulate the approach was developed by Matlab. One noisy life-time IR images are used as test samples to assess the effect on the proposed approach. Their details are described in the following section.

4.2 Test samples of the IR image

In order to validate the proposed approach, one gray-scale noisy life-time IR images are collected as test sample, as shown in Fig. 2. There are 364 × 244 pixels in the images, respectively, and each pixel is represented by 8-bits in gray-scale. Fig. 2 shows people walking on the street on a rainy day. The people (objects) are small compared with the street scenery rendered as background.

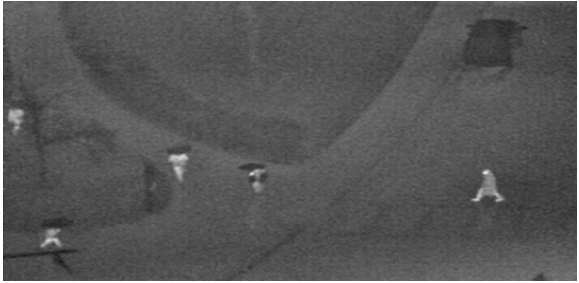

Fig. 2 An Infrared image taken of some people walking on the street on a rainy day.

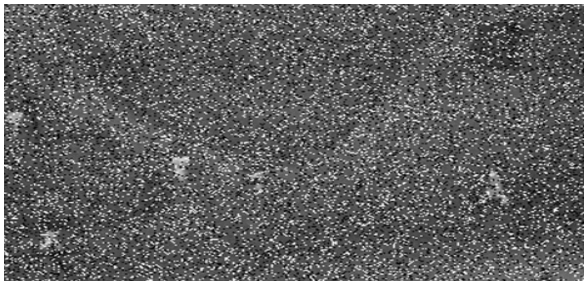

Fig. 3 Shows Fig. 2 with impulse noise processed by FNR iteratively with 20% impulse noise.

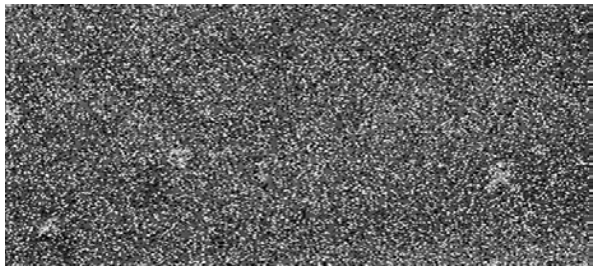

Fig. 4 Shows Fig. 2 with impulse noise processed by FNR iteratively with 30% impulse noise.

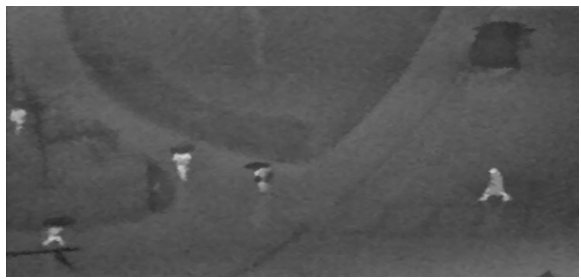

Fig. 5 The result of Fig. 3. iteratively processed by Fast noise reduction approach two times.

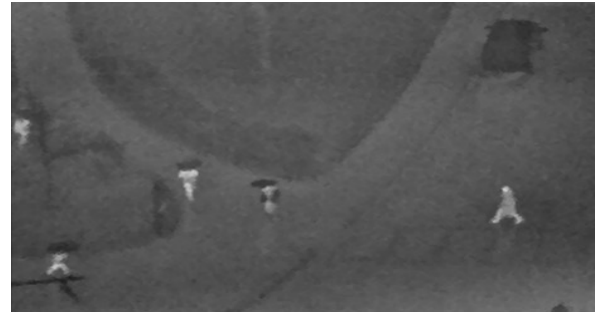

Fig. 6 The result of Fig. 4. iteratively processed by Fast noise reduction approach two times.

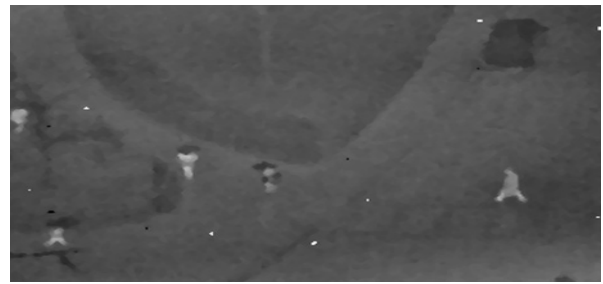

Fig. 7 The result of Fig.3. iteratively processed by MF two times,.

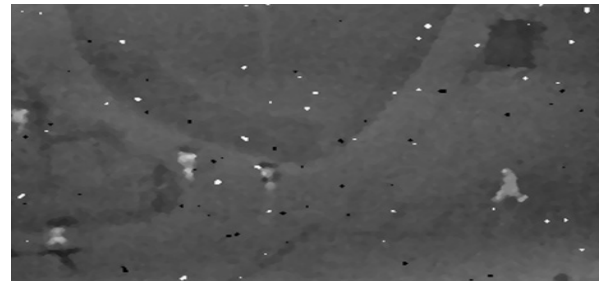

Fig. 8 The result of Fig.4. iteratively processed by MF two times,.

In order for the small objects of interest in these IR images to be observed easily, quickly, and accurately, they have to be preprocessed using a noise reduction.

The performances of FNR approach are address in this paragraph. Fig. 3 & 4 show the Fig. 2 with impulse noise 20%, 30%. Fig. 5 & 6 exhibit the result of performing FNR to iteratively filter two times. the impulse noise is filtered out and preserve the edges, textures and detail information simultaneously Fig.7 & 8 iteratively processed by MF two times they show the more noises are not filtered out and more edges, textures and detail information are lost. In addition, the proposed FNR approach can preserve the edge and texture information regardless of increases in filter size while removing noise. When the filter size increases, the image processed by the MF is blurred, and the edges and texture information are lost. The experimental results of performing the MF and FNR Table 1 illustrates the different performances between utilizing the MF and FNR approach on Fig. 3 and 4 one time. The MF processes each pixel in a whole image so it has to perform a sorting algorithm 88,816 times for Fig. 3 and spend 24 ms. On the other hand, the FNR approach performs a sorting algorithm for noisy pixels only, performing the sorting algorithm 24,587 times for Fig. 3 and spending 15 ms, which includes finding the $g_{max}$ and $g_{min}$ 88,816 times. Hence, the computation load is reduced 37.5% by the proposed FNR approach. The FNR reduces the computation load 25%.

A typical noise measure used is peak signal-to-noise ratio (PSNR) [3], defined as.

$$PSNR = 10 \log_{10}(MAX^2 / MSE) \qquad (5)$$

where MSE is the mean square error between the original and processed image and Max is the maximum gray scale of pixels, e.g., 255 for 8 bits. We use the PSNR to assess the noise reduction performance of the FNR approach and MF.

Table 1 : The differences in performance between utilizing the MF and the FNR approach on Fig. 3&4.

| Image Size (Pixels) | 364×244 (88, 816) | | |
|---|---|---|---|
| Noise Image | | Fig.3(20% noise) | Fig.4(30% noise) |
| Noisy image | PSNR(db) | 34.11 | 32.26 |
| Processed by MF | Sort times | 88,816 | 88,816 |
| | Process time(ms) | 24 | 24 |
| | PSNR(db) | 40.23 | 38.79 |
| Processed by FNR approach | Time of finding $g_{max}$ & $g_{min}$ | 88,816 | 88,816 |
| | Sort times | 24,587 | 30,832 |
| | Process Time(ms) | 15 | 18 |
| | PSNR(db) | 43.45 | 41.39 |
| Process time improved by FNR (%) | | 37.50% | 25% |
| PSNR improved by FNR (dB) | | 3.22 | 2.60 |

## 5. Conclusions

In this paper, we present an effective FNR approach to reduce the noise of IR images. Based on the results shown in Figs. 5,6,7 and 8 and in Tables 1 the FNR approach possesses three main advantages. The first is that FNR approach utilizes noise detection based on IR imaging mechanism to identify noisy pixels in IR images, and median-based noise removal performed by Intro sorting with bits decomposition effectively decreases the computation load for performing noise reduction. It can be applied to real-time video processing by software. The second advantage of FNR approach is that remedies the shortcomings of MF while increasing the filter window size. Finally, no prior knowledge about the IR images. necessary and no parameter must be manually preset to perform the proposed approach.

Experimental results demonstrate that the proposed approach can improve the performance of noise reduction and enhance quality IR images. In the applications of IR images, it is a considerable challenge to provide a removal on noise but not on the edges, text information and small objects of interest. In order to conquer the challenge, we propose an FNR approach consisting of the noise detection and noise removal. Experimental results demonstrate that the proposed approach can meet this challenge.

### Acknowledgments

This work is supported by National Natural Science Foundation , resources from Prof. Dr. M. Rizwan Beg, Head of department of computer science and engineering, Integral University, Lucknow. They would also thank the anonymous reviewers for their significant and constructive critiques and suggestions, which substantially improved the quality of this paper.

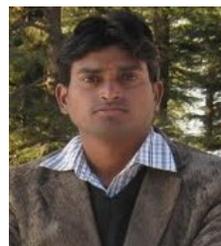

**Kapil Kumar Gupta** obtained his B. Tech (IT) degree in 2009 from JSS academy of technical education, Noida, Uttar Pradesh, India . He is currently a student of M. Tech ( CSE) from Integral University, Lucknow and Astt. Professor in Department of Information Technology, Goel Institute of technology and management, Lucknow. Uttar Pradesh, India. His main research interest is in the field of Image Processing, Design and analysis of algorithms, etc.. He is a  Member of CSTA.

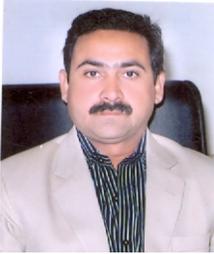**Prof. Dr. M. Rizwan Beg** is M. Tech & Ph.D in Computer Sc. & Engg. Presently he is working as Professor & Head Deptt. Of Computer Sc. & Engg. and Information Technology of Integral University Lucknow, Uttar Pradesh, India. He is having more than 16 years of experience which includes around 14 years of teaching experience. His area of expertise is Software Engg., Requirement Engineering, Software Quality, and Software Project Management. He has published more than 40 Research papers in International Journals & Conferences. Presently 8 research scholars are pursuing their Ph.D in his supervision. Dr. Beg is Associate Editor in Chief of Advancement in Computing Technology & is also member of editorial board for various other International & National Journals. He is member of large member of International professional societies & also member of Advisory Board for various institutions in India. He chaired a no. of workshops, seminars & National Conference.

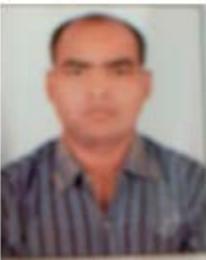**Jitendra Kumar Niranjan** has obtained his M.Tech (Information System) degree in 2011 from Delhi Technological University, Delhi, India. He is currently working as Astt. Professor in IMS Engineering College Ghaziabad. His main research interest is in the field of Image Processing, and Cloud computing.